\documentclass[12pt]{article}
\usepackage{a4wide}
\usepackage[english]{babel}
\usepackage{graphicx}
\usepackage[font=small,labelfont=bf]{caption}
\usepackage{amsfonts}
\usepackage{amsmath}
\usepackage{amsthm}
\usepackage{algorithm}
\usepackage{algpseudocode}
\usepackage{bbm}
\usepackage{booktabs} 
\usepackage{authblk}
\usepackage{subfigure}
\usepackage{amsthm}
\usepackage{natbib}
\usepackage{hyperref}
\usepackage{fancyhdr}
\usepackage[T1]{fontenc}
\usepackage[utf8]{inputenc}
\usepackage{authblk}
\hypersetup{colorlinks   = true,
                   citecolor = blue, 
                   linkcolor = blue}
\usepackage[width=0.88\textwidth]{caption}
\usepackage[titletoc,title]{appendix}

\title{Sparse multi-view matrix factorisation: a multivariate approach to multiple tissue comparisons}
\author[1]{Zi Wang}
\author[2]{Wei Yuan}
\author[1,3]{Giovanni Montana}
\affil[1]{Department of Mathematics, Imperial College London, London SW7 2AZ, UK.}
\affil[2]{Department of Twin Research and Genetic Epidemiology, King's College London, St Thomas' Hospital, SE1 7EH, UK.}
\affil[3]{Department of Biomedical Engineering, King's College London, St Thomas' Hospital, London SE1 7EH, UK.}

\date{}

\begin{document}

\maketitle

\abstract{Within any given tissue, gene expression levels can vary extensively among individuals. Such heterogeneity can be caused by genetic and epigenetic variability and may contribute to disease. The abundance of experimental data now enables the identification of features of gene expression profiles that are shared across tissues and those that are tissue-specific. While most current research is concerned with characterizing differential expression by comparing mean expression profiles across tissues, it is believed that a significant difference in a gene expression's variance across tissues may also be associated with molecular mechanisms that are important for tissue development and function. 

We propose a sparse multi-view matrix factorization (sMVMF) algorithm to jointly analyse gene expression measurements in multiple tissues, where each tissue provides a different `view' of the underlying organism. The proposed methodology can be interpreted as an extension of principal component analysis in that it provides the means to decompose the total sample variance in each tissue into the sum of two components: one capturing the variance that is shared across tissues and one isolating the tissue-specific variances. sMVMF has been used to jointly model mRNA expression profiles in three tissues obtained from a large and well-phenotyped twins cohort, TwinsUK. Using sMVMF, we are able to prioritize genes based on whether their variation patterns are specific to each tissue. Furthermore, using DNA methylation profiles available, we provide supporting evidence that adipose-specific gene expression patterns may be driven by epigenetic effects.
}

\section{Introduction}  \label{introduction}

RNA abundance, as the results of active gene expression, affects cell differentiation and tissue development \citep{coulon13}. As such, it provides a snapshot of the undergoing biological process within certain cells or a tissue. Except for house-keeping genes, the expressions of a large number of genes vary from tissue to tissue, and some may only be expressed in a particular tissue or a certain cell type  \citep{xia07}. The regulation of tissue-specific expression is a complex process in which a gene's enhancer plays a key role regulating gene expressions via DNA methylation \citep{ong11}. Genes displaying tissue-specific expressions are widely associated with cell type diversity and tissue development \citep{reik07}, and aberrant tissue-specific expressions have been associated with diseases that originated in the underlying tissue \citep{veer02, lage08}. 
Distinguishing tissue-specific expressions from expression patterns prevalent in all tissues holds the promise to enhance fundamental understanding of the universality and specialization of molecular biological mechanisms, and potentially suggest candidate genes that may regulate traits of interest \citep{xia07}.  As collecting genome-wide transcriptomic profiles from many different tissues of a given individual is becoming more affordable, large population-based studies are being carried out to compare gene expression patterns across human tissues \citep{liux08,yang11}.

A common approach to detecting tissue-specific expressions consists of comparing the \emph{mean} expression levels of individual genes across tissues. This can be accomplished using standard univariate test statistics. For instance, \cite{wu14} used the two-sample Z-test to compare non-coding RNA expressions in three embryonic mouse tissues: they reported approximately 80\% of validated \emph{in vivo} enhancers exhibited tissue-specific RNA expression that correlated with tissue-specific enhancer activity. \cite{yang11} applied a modified version of Tukey's range test \citep{tukey49}, a test statistic based on the standardised mean difference between two groups, to compare expression levels of 127 human tissues, and results of this study are publicly available in the VeryGene database. A related database, TiGER \citep{liux08}, has also been created by comparing expression sequence tags (EST) in 30 human tissues using a binomial test on EST counts. Both VeryGene and TiGER contain up-to-date annotated lists of tissue-specific gene expressions, which generated hypotheses for studies in the area of pathogenic mechanism, diagnosis, and therapeutic research \citep{wu09}. 

More recent studies have gone beyond the single-gene comparison and aimed at extracting multivariate patterns of differential gene expression across tissues. \cite{xiao14} applied the higher-order generalised singular value decomposition (HO-GSVD) method proposed by \cite{ponnapalli11} and compared co-expression networks from multiple tissues. This technique is able to highlight co-expression patterns that are equally significant in all tissues or exclusively significant in a particular tissue. The rationale for a multivariate approach is that when a gene regulator is switched on, it can raise the expression level of all its downstream genes in specific tissues. Hence a multi-gene analysis may be a more powerful approach. 

While most studies explore the differences in the mean of expression, the sample variance is another interesting feature to consider. Traditionally, comparison of expression variances has been carried out in case-control studies \citep{mar11}. Using an F-test, significantly high or low gene expression variance has been observed in many disease populations including lung adenocarcinoma and colerectal cancer, whereas the difference in mean expression levels was not found significant between cases and controls \citep{ho08}. In a tissue-related study, \cite{cheung03} carried out a genome-wide assessment of gene expressions in human lymphoblastoid cells. Using an F-test, the authors showed that high-variance genes were mostly associated with functions such as cytoskeleton, protein modification and transport, whereas low-variance genes were mostly associated with signal transduction and cell death/proliferation. 

In this work we introduce a novel multivariate methodology that can detect patterns of differential variance across tissues. We regard the gene expression profiles in each tissue as providing a different ``view'' of the underlying organism and propose an approach to carry out such a multi-view analysis. Our objective is to identify genes that jointly explain the same amount of sample variance in all tissues - the "shared" variance - and genes that explain substantially higher variances in each specific tissue separately - the "tissue-specific" variances - while the shared variance has been accounted for. During this process we impose a constraint that the factors driving shared and tissue-specific variability must be uncorrelated so that the total sample variance can be decomposed into the two corresponding components. The proposed methodology, called sparse multi-view matrix factorisation (sMVMF), can be interpreted as an extension of principal component analysis (PCA), which is traditionally used to identify a handful of latent factors explaining a large portion of sample variance separately in each tissue.

The rest of this paper is organised as follows. The sMVMF methodology is presented in Section \ref{sMVMF:methods},where we also discuss connections with a traditional PCA and derive the parameter estimation algorithm. In Sectionv \ref{sMVMF:simulation} we demonstrate the main feature of the proposed method on simulated data, and report on comparison with alternative univariate and multivariate approaches.  In Section \ref{sMVMF:application} we apply the sMVMF to compare mRNA expressions in three tissues obtained from a large twin population, the TwinsUK cohort. We conclude in Section {\ref{sMVMF:discussion} with a discussion.

\section{Methods} \label{sMVMF:methods}

\subsection{Sparse multi-view matrix factorisation}  \label{sMVMF:model}

We assume to have collected $p$ gene expression measurements for $M$ different tissues. Ideally the data for all tissues should be derived from the same underlying random sample (as in our application, Section \ref{sMVMF:application}) in order to remove sources of biological variability that can potentially induce differences in gene expression profiles across tissues. In practice, however, cross-tissue experiments rarely collect samples from the same set of subjects or may fail quality control. In our setting therefore we assume $M$ different random samples, each one contributing a different tissue dataset. The $m^{th}$ dataset consists of $n_m$ subjects, and the expression profiles are arranged in an $n_m \times p$ matrix. All} matrices are collected in $\mathcal{X} = \{X^{(1)}$, $X^{(2)}$, ..., $X^{(M)}\}$, where the superscripts refer to tissue indices. For each $X^{(m)}$, we subtract the column mean from each column such that each diagonal entry of the scaled gram matrix, $\frac{1}{n_m}(X^{(m)})^TX^{(m)}$, is proportional to the sample variance of the corresponding variable, and the trace is the total sample variance. We aim to identify genes that jointly explain a large amount of sample expression variances in all tissues and genes that explain substantially higher variances in a specific tissue. Our strategy involves approximating each ${\color{black}\frac{1}{\sqrt{n_m}}}X^{(m)}$ by the sum of a shared variance component and a tissue-specific component:
\begin{equation}  \label{sMVMF:prototype}
\frac{1}{\sqrt{n_m}}X^{(m)}~~~  \approx \underbrace{S^{(m)}}_\text{shared variance component} + ~~~ \underbrace{T^{(m)}}_\text{tissue-specific variance component} 
\end{equation}
for $m=1,2,...,M$, where $1/\sqrt{n_m}$ is a scaling factor such that the trace of the gram matrix of the left-hand-side equals the sample variance. These components are defined so as to yield the following properties: 
\begin{itemize}
\item[(a)] The rank of $S^{(m)}$ and $T^{(m)}$ are both much smaller than min$(n_m,p)$  so that the two components provide insights into the intrinsic structure of the data while discarding redundant information. 
\item[(b)] The variation patterns captured by shared component are uncorrelated to the variation patterns captured by tissue-specific component. As a consequence of this, the total variance explained by $S^{(m)}$ and $T^{(m)}$ altogether equals the sum of the variance explained by each individual component. 
\item[(c)] The shared component explains the same amount of variance of each gene expression in all tissues. As such, the difference in expression variance between tissues is exclusively captured in tissue-specific variance component.

\end{itemize}

We start by proposing a factorisation of both $S^{(m)}$ and $T^{(m)}$ which, by imposing certain constraints, will satisfy the above properties.  Suppose rank$(S^{(m)})=d$ and rank$(T^{(m)})=r$, where $d,r << \text{min}(n_m,p)$ following property (a). For a given $r$, $T^{(m)}$ can be expressed as the product of an $n_m \times r$ full rank matrix $W^{(m)}$ and the transpose of a $p \times r$ full rank matrix $V^{(m)}$, that is:
\begin{equation}   \label{sMVMF:Tfactorisation}
T^{(m)}=W^{(m)}(V^{(m)})^T=\sum_{j=1}^{r}W^{(m)}_j (V^{(m)}_j)^T=\sum_{j=1}^{r}T^{(m)}_{[j]}
\end{equation}
where the superscript $^T$ denotes matrix transpose, and the subscript $j$ denotes the $j^{th}$ column of the corresponding matrix. Each 
$$T^{(m)}_{[j]}:=W^{(m)}_j (V^{(m)}_j)^T$$
has the same dimension as $T^{(m)}$ and is composed of a tissue-specific latent factor (LF). A LF is an unobservable variable assumed to control the patterns of observed variables and hence may provide insights into the intrinsic mechanism that drives the difference of expression variability between tissues. The matrix factorisation in \eqref{sMVMF:Tfactorisation} is not unique, since for any $r \times r$ non-singular square matrix $R$, $T^{(m)}=W^{(m)}(V^{(m)})^T=(W^{(m)}R)(R^{-1}(V^{(m)})^T)=\tilde{W}^{(m)}(\tilde{V}^{(m)})^T$. We introduce an orthogonal constraint \\ $(W^{(m)})^TW^{(m)}=I_r$ so that the matrix factorisation is unique subject to an isometric transformation. Similarly, we can factorise the shared component as:
\begin{equation}    \label{sMVMF:Sfactorisation}
S^{(m)}=U^{(m)}(V^*)^T=\sum_{k=1}^{d}U^{(m)}_k (V^*_k)^T=\sum_{k=1}^{d}S^{(m)}_{[k]}  
\end{equation} 
where $U^{(m)}$ is orthogonal and $V^*$ is tissue-independent which we shall explain. Each $S^{(m)}_{[k]}$ has the same dimension as $S^{(m)}$ and is composed of one shared variability LF. The resulting multi-view matrix factorisation (MVMF) then is:
\begin{equation}  \label{sMVMF:mvpcadecomp}
\frac{1}{\sqrt{n_m}}X^{(m)} ~~ \approx ~ U^{(m)}(V^*)^{T}+W^{(m)}(V^{(m)})^{T}
\end{equation}

The matrix factorisations \eqref{sMVMF:Tfactorisation} and \eqref{sMVMF:Sfactorisation} are intimately related to the singular value decomposition (SVD) of $S^{(m)}$ and $T^{(m)}$. Specifically, $U^{(m)}$ and $W^{(m)}$ are analogous to the matrix of left singular vectors and also the principal components (PCs) in a standard PCA. They represent gene expression patterns in a low-dimensional space where each dimension is derived from the original gene expression measurements such that the maximal amount of variance is explained. We shall refer the columns of $U^{(m)}$ and $W^{(m)}$ as the principal projections (PPJ). $(V^*)^T$ and $(V^{(m)})^T$ are analogous to the product of the diagonal matrix of eigenvalues and the matrix of right singular vectors. Since the singular values determine the amount of variance explained and the right singular vectors correspond to the loadings in the PCA which quantifies the importance of the genes to the expression variance explained, using the same matrix $V^*$ for all tissues in the shared component results in the same amount of shared variability explained for each gene expression probe, such that property (c) is satisfied. We shall refer to matrices $V^*$ and $V^{(m)}$ as transformation matrices. 

A sufficient condition to satisfy property (b) is:
\begin{equation}    \label{sMVMF:uworth}
(U^{(m)})^TW^{(m)}=0_{d \times r}
\end{equation}
This constraint, in addition to the orthogonality of $U^{(m)}$ and $W^{(m)}$, results in the $(d+r)$ PPJs represented by $[U^{(m)}, W^{(m)}]$ being pairwise orthogonal, which is analogous to the standard PCA where the PCs are orthogonal. Intuitively, this means for each tissue the LFs driving shared and tissue-specific variability are uncorrelated. The amount of variance explained in tissue $m$, $\hat{\sigma}_m^s$, can be computed as (subject to a constant factor):
\begin{equation}    \label{sMVMF:TotalVar} \hat{\sigma}_m^s=\text{Tr}\{(S^{(m)})^TS^{(m)}+(T^{(m)})^TT^{(m)}+2(S^{(m)})^TT^{(m)}\}
\end{equation}
where $\emph{Tr}$ denotes the matrix trace.  Recalling that $S^{(m)}=U^{(m)}(V^*)^T$ and $(U^{(m)})^TU^{(m)}=I_d$, the amount of shared variance explained is:
\begin{equation}    \label{sMVMF:SharedVar}
\sigma_*=\text{Tr}\{(S^{(m)})^TS^{(m)}\} = \text{Tr}\{V^*(V^*)^T\}
\end{equation}
Likewise, recalling that $T^{(m)}=W^{(m)}(V^{(m)})^T$ and $(W^{(m)})^TW^{(m)}=I_r$, the amount of tissue-specific variance explained is:
\begin{equation}    \label{sMVMF:SpecificVar}
\sigma_m=\text{Tr}\{(T^{(m)})^TT^{(m)}\}=\text{Tr}\{V^{(m)}(V^{(m)})^T\}
\end{equation}
Making the same substitutions into \eqref{sMVMF:TotalVar}, we obtain:
\begin{eqnarray}   
\hat{\sigma}_m^s = \text{Tr}\{V^*(V^*)^T+V^{(m)}(V^{(m)})^T+2V^*(U^{(m)})^TW^{(m)}(V^{(m)})^T\}    \nonumber
\end{eqnarray}
Substituting \eqref{sMVMF:uworth} into the above equation, we reach:
\begin{equation}    \label{sMVMF:vardecomp}
\hat{\sigma}_m^s = \text{Tr}\{V^*(V^*)^T+V^{(m)}(V^{(m)})^T\} = \sigma_*+\sigma_m
\end{equation}
which satisfies (b).

\subsection{Sparsity constraints and estimation}  \label{sMVMF:estimation}

The factorisation \eqref{sMVMF:mvpcadecomp} is obtained by minimising the squared error. This amounts to minimising the loss function:
\begin{equation}    \label{sMVMF:mvpcaloss}
\ell = \sum_{m=1}^{M} \Arrowvert \frac{1}{\sqrt{n_m}}X^{(m)} - U^{(m)}(V^*)^{T} - W^{(m)}(V^{(m)})^{T} \Arrowvert_\mathcal{F}^2
\end{equation}
where $\Arrowvert . \Arrowvert_\mathcal{F}$ refers to the Frobenius norm, subject to the following orthogonality constraints:
\begin{eqnarray}    \label{sMVMF:mvpcaconstraint}
(U^{(m)})^{T}U^{(m)} = I, ~ (W^{(m)})^{T}W^{(m)} = I, ~ (U^{(m)})^{T}W^{(m)} = 0.
\end{eqnarray}
For fixed $U^{(m)}(V^*)^{T}$, the optimal $T^{(m)}=W^{(m)}(V^{(m)})^T$ is a low-rank approximation of $\Arrowvert \frac{1}{\sqrt{n_m}}X^{(m)} - S^{(m)} \Arrowvert_\mathcal{F}^2$, where each rank sequentially captures the maximal variance remained in each data matrix after removing the shared variability. Likewise, for fixed $W^{(m)}(V^{(m)})^{T}$, each rank of the optimal $S^{(m)}=U^{(m)}(V^*)^{T}$ sequentially captures the maximal variance remained across all tissues after removing the tissue-specific variance.

In transcriptomics studies, it is widely believed that the differences in gene expressions between cell and tissue types are largely determined by transcripts derived from a small number of tissue-specific genes \citep{jongeneel05}. Therefore it seems reasonable that in our application of multi-tissue comparison of gene expressions, for each PPJ, the corresponding column in the transformation matrix should feature a limited number of non-zero entries. In such a scenario, a sparse representation will not only generate more reliable statistical models by excluding noise features, but also offer more biological insight into the underlying cellular mechanism \citep{ma08}. 

In the context of MVMF, we induce sparse estimates of $V^*$ and $V^{(m)}$ by adding penalty terms to the loss function $\ell~(U,W,V^*,V)$ as in \eqref{sMVMF:mvpcaloss}. Specifically, we minimise:
\begin{equation}    \label{sMVMF:mvpcaobjective}
\ell~(U,W, V^*,V) + 2 \cdot M \cdot \Arrowvert V^* \Lambda^* \Arrowvert_1 + 2 \sum_{m=1}^{M} \Arrowvert V^{(m)} \Lambda^{(m)} \Arrowvert_1
\end{equation}
where $\Arrowvert ~ \Arrowvert_1$ denotes the $\ell_1$ norm. $\Lambda^*$ and $\Lambda^{(m)}$ are $d \times d$ and $r \times r$ diagonal matrices, respectively. In both matrices, the $k^{th}$ diagonal entry is a non-negative regularisation parameter for the $k^{th}$ column of the corresponding transformation matrix, and the $k^{th}$ column tends to have more zero entries as the $k^{th}$ diagonal entry increases. In practice, a parsimonious parametrisation may be employed where $\Lambda^*=\lambda_1I_d$ and $\Lambda^{(m)}=\lambda_2 I_r$ for $m=1,...,M$ so that the number of parameters to be specified is greatly reduced. Alternatively, $\Lambda^*$ and $\Lambda^{(m)}$ may be set such that a specified number of variables are selected in each column of $\hat{V}^*$ and $\hat{V}^{(m)}$.

The optimisation problem \eqref{sMVMF:mvpcaobjective} with constraints \eqref{sMVMF:mvpcaconstraint} is not jointly convex in $U^{(m)}$, $W^{(m)}$, $V^{(m)}$, and $V^*$ for $m=1,2,...,M$ (for instance the orthogonality constraints are non-convex in nature), hence gradient descent algorithms will suffer from multiple local minima \citep{gorski07}. We propose to solve the optimisation problem by alternately minimising with respect to one parameter in $U^{(m)}$,$W^{(m)}$,$V^*$, $V^{(m)}$ while fixing all remaining parameters, and repeating this procedure until the algorithm converges numerically. The minimisation problem with respect to $V^*$ or $V^{(m)}$ alone is strictly convex, hence in these steps a coordinate descent algorithm (CDA) is guaranteed to converge to the global minimum \citep{friedman07}. CDA iteratively update the parameter vector by cyclically updating one component of the vector at a time, until convergence. On the other hand, the minimisation problem with respect to $W^{(m)}$ or $U^{(m)}$ is not convex. For fixed $V^*$ and $V^{(m)}$, the estimates of $W^{(m)}$ and $U^{(m)}$ that minimise \eqref{sMVMF:mvpcaobjective} can be jointly computed via a closed form solution. Assuming we have obtained initial estimates of $V^*$ and $V^{(m)}$, we cyclically update the parameters in the following order:
$$(U^{(m)}, W^{(m)}) \rightarrow V^{(m)} \rightarrow V^*$$ 
Here $U^{(m)}$ and $W^{(m)}$ are jointly estimated in the first step, and in the subsequent steps $V^{(m)}$ and $V^*$ are updated separately, while keeping the previous estimates fixed. A detailed explanation of how each update is performed is in order.

First we reformulate the estimation problem as follows: we bind the columns of $U^{(m)}$ and $W^{(m)}$ and define the $n_m \times (d+r)$ augmented matrix: $\tilde{U}^{(m)} = [U^{(m)} ~,~ W^{(m)}]$; we then bind the columns of $V^*$ and $V^{(m)}$ and define the $p\times (d+r)$ matrix: $\tilde{V}^{(m)} = [V^* ~,~ V^{(m)}]$. As such:
\begin{eqnarray}
\ell~(U,W,V^*,V^{(m)})=\sum_{m=1}^{M} \Arrowvert \frac{1}{\sqrt{n_m}}X^{(m)} - \tilde{U}^{(m)}(\tilde{V}^{(m)})^{T} \Arrowvert_\mathcal{F}^2    \nonumber
\end{eqnarray}
and the constraints in \eqref{sMVMF:mvpcaconstraint} can be combined into:
\begin{eqnarray}
(\tilde{U}^{(m)})^T\tilde{U}^{(m)} = I_{d+r}    \nonumber
\end{eqnarray}
Fixing $\tilde{V}^{(m)}$, the estimate of $\tilde{U}^{(m)}$ can be obtained by the reduced-rank Procrustes rotation procedure which seeks the optimum rotation of $X^{(m)}$ such that the error $\Arrowvert \frac{1}{\sqrt{n_m}}X^{(m)} - \tilde{U}^{(m)}(\tilde{V}^{(m)})^{T} \Arrowvert_\mathcal{F}^2$ is minimal. For a proof of this, see \citep{zou06spca}. We obtain the SVD of $\frac{1}{\sqrt{n_m}}X^{(m)}\tilde{V}^{(m)}$ as $PQR^T$, and compute the estimate of $\tilde{U}^{(m)}$ by: $\hat{\tilde{U}}^{(m)}=PR^T$. 


Next, we fix $U^{(m)}$, $W^{(m)}$, and $V^*$ while minimising \eqref{sMVMF:mvpcaobjective} with respect to $V^{(m)}$. For each fixed $m$, varying $V^{(m)}$ only changes the objective function via the summand indexed $(m)$. Hence it is sufficient to minimise:
\begin{equation}    \label{sMVMF:vobj}
\Arrowvert \frac{1}{\sqrt{n_m}}X^{(m)} - U^{(m)}(V^*)^T - W^{(m)}(V^{(m)})^T \Arrowvert_\mathcal{F}^2 + 2 \Arrowvert V^{(m)} \Lambda^{(m)} \Arrowvert_1 .
\end{equation}
This function is strictly convex in $V^{(m)}$ and the CDA is guaranteed to converge to the global minimum. We drop the superscript $(m)$ in the following derivation for convenience and denote the $j^{th}$ column of the matrix $V$ by $V_j$. In each iteration, the estimate of $V_j$ is found by equating the first derivative of \eqref{sMVMF:vobj} with respect to $V_j$ to zero. Hence:
\begin{eqnarray}
-2(\frac{1}{\sqrt{n_m}}X-UV^*-WV^T)^TW_j + 2 \Lambda_j \cdot \nabla(|V_j|) = 0,  \nonumber
\end{eqnarray}
where $\nabla$ is the gradient operator. Substitute \eqref{sMVMF:mvpcaconstraint} and rearrange to give:
\begin{eqnarray}
V_j = \frac{1}{\sqrt{n_m}}X^TW_j - \Lambda_j \cdot \nabla(|V_j|)     \nonumber
\end{eqnarray}
We define the sign function $\sigma(y)$ which equals $1$ if $y>0$, $-1$ if $y<0$, and 0 if $y=0$. First note the derivative of the function $|y|$ is $\sigma(y)$ if $y \ne 0$ and a real number in the interval $(-1,1)$ otherwise. Rearrange the previous equation to obtain the updated estimate in each iteration:
\begin{equation}    \label{sMVMF:vupdateeq}
\hat{V}^{(m)}_j = S_{\Lambda^{(m)}_j}\big( (\frac{1}{\sqrt{n_m}}X^{(m)})^T W^{(m)}_j \big)
\end{equation}
where $S_\lambda(y)$ is a soft-thresholding function on vector $y$ with non-negative parameter $\lambda$ such that $S_\lambda(y) = \sigma(y) \cdot \text{max} \{ |y|-\lambda, 0\}$, and $\Lambda^{(m)}_j$ is the $j^{th}$ diagonal entry of $\Lambda^{(m)}$.

In the third step, we fix the estimates of $U^{(m)}$, $W^{(m)}$, and $V^{(m)}$ and minimise \eqref{sMVMF:mvpcaobjective} with respect to $V^*$. The objective function becomes:
\begin{equation}    \label{sMVMF:vstarobj}
\ell + 2 \cdot M \cdot \Arrowvert V^* \Lambda^* \Arrowvert_1
\end{equation}
where $\ell$ is defined in \eqref{sMVMF:mvpcaloss}. As in the second step, we use a CDA in each iteration and the updated estimate of $V^*_i$ is found by equating the first derivative of \eqref{sMVMF:vstarobj} to zero. Specifically: 
\begin{eqnarray}
&-2\sum_{m=1}^{M} \bigg\{[\frac{1}{\sqrt{n_m}}X^{(m)}-U^{(m)}V^*-W^{(m)}(V^{(m)})^T]^T U_i \bigg\}  \nonumber \\
&+~2 \cdot M \cdot \Lambda^*_i  \cdot \nabla(|V^*_i|) = 0,  \nonumber
\end{eqnarray}
where $\Lambda^*_i$ is the $i^{th}$ diagonal entry of $\Lambda^*$. Applying \eqref{sMVMF:mvpcaconstraint}, this can be re-arranged into:
\begin{eqnarray}   
M \cdot V^*_i = \sum_{m=1}^{M} (\frac{1}{\sqrt{n_m}}X^{(m)})^TU^{(m)}_i - M \cdot \Lambda^*_i \cdot \nabla(|V^*_i|),    \nonumber
\end{eqnarray}
Using the soft-thresholding and the sign functions, the updated estimate in each iteration can be re-written as:
\begin{equation}    \label{sMVMF:vstarupdateeq}
\hat{V}^*_i = S_{\Lambda^*_i} \bigg( \frac{1}{M} \sum_{m=1}^{M} (\frac{1}{\sqrt{n_m}}X^{(m)})^TU^{(m)}_i \bigg)
\end{equation}
The cyclic CDA requires initial estimates of $V^*$ and $V^{(m)}$, which are obtained as follows. First we set an initial value to $V^*$, which explains as much variance in all datasets in $\mathcal{X}$ as possible. This amounts to a PCA on the $(\sum_{m=1}^M n_m) \times p$ matrix $\check{X}$ obtained by binding the rows of $\frac{1}{\sqrt{n_m}}X^{(m)}$, $m=1,...,M$. We compute the truncated SVD of $\check{X}$ and obtain $\check{X}=\check{U}DB^T$ where $D$ contains the $d$ largest eigenvalues of $\check{X}^T\check{X}$. The initial estimate of $V^*$ is then defined as:
\begin{equation}    \label{sMVMF:initialVstar}
(\hat{V}^*)^T=\frac{1}{M}DB^T,
\end{equation}
and $\hat{U}^{(m)}$ is defined by the corresponding rows of $\check{U}$ in the SVD. For the tissue-specific transformation matrices $V^{(m)}$, we compute the SVD of the residuals after removing the shared variance component from $\frac{1}{\sqrt{n_m}}X^{(m)}$, which gives: $\frac{1}{\sqrt{n_m}}X^{(m)} - \hat{U}^{(m)}\hat{V}^* = W^{(m)}R^{(m)}(Q^{(m)})^T$. The initial estimate of $V^{(m)}$ is defined as:
\begin{equation}    \label{sMVMF:initialV}
(\hat{V}^{(m)})^T = R^{(m)}(Q^{(m)})^T.   
\end{equation} 
A summary of the estimation procedure is given in Algorithm ~\ref{sMVMF:alg:sMVMF}. 

\begin{algorithm}
\caption{sMVMF estimation algorithm}
\label{sMVMF:alg:sMVMF}
$\mathbf{Input}$: data $\mathcal{X}$; parameters $d$, $r$, $\Lambda^{(m)}$, $\Lambda^*$ for $m=1,2,...,M$.\\
$\mathbf{Output}$: $U^{(m)}$, $W^{(m)}$, $V^{(m)}$, for $m=1,2,...,M$, and $V^*$.
\begin{algorithmic}[1]
\State Get initial estimates of $V^{(m)}$ for $m=1,2,...,M$, and $V^*$ as in \eqref{sMVMF:initialV} and \eqref{sMVMF:initialVstar}.
\State $\mathbf{while}$ not convergent $\mathbf{do}$:
\State ~~~ Apply SVD: $\frac{1}{\sqrt{n_m}}X^{(m)}\tilde{V}^{(m)} = PQR^T$, and set $\hat{\tilde{U}}^{(m)}=PR^T$.
\State ~~~ Use CDA to estimate $V^{(m)}$ according to \eqref{sMVMF:vupdateeq}.
\State ~~~ Use CDA to estimate $V^*$ according to corollary \eqref{sMVMF:vstarupdateeq}.
\end{algorithmic}
\end{algorithm}

\subsection{Parameter selection}    \label{sMVMF:tuning}

The sMVMF contains two sets of parameters: the tissue-specific sparsity parameters $\Lambda^{(m)}$, $\Lambda^*$, and the $(d,r)$ pair. Both $d$ and $r$ balance model complexity and the amount of variance explained. We select the smallest possible values of $d$ and $r$ such that a prescribed proportion of variance is explained. For a fixed $(d,r)$ pair, the sparsity parameters can be optimised using a cross-validation procedure, which identifies the best combination from a grid of candidate values so that the amount of variance explained is maximised on the testing data for the chosen $(d,r)$. However, in high-dimensional settings, cross-validation procedures such as this one tend to favour over-complex models which may include noise variables \citep{hdbook2011}. Instead we propose using the ``stability selection'' procedure which is particularly effective in improving variable selection accuracy and reducing the number of false positives in high-dimensional settings \citep{meinshausen10}. Given parameters $d=d_0$ and $r=r_0$ in sMVMF, variables can be ranked according to their importance in explaining shared and tissue-specific variances by applying a stability selection procedure as follows:

\begin{enumerate}
\item Randomly extract half of the $n_m$ samples from each $X^{(m)}$ without replacement and denote the resulting data matrix $X^{(m)}_s$, for $m=1,...,M$. In the case where each $X^{(m)}$ consists of the same subjects, it may be preferable to draw the same samples from all datasets. 

\item Fit sMVMF on $X^{(m)}_s$, $m=1,...,M$, where $\Lambda^*$ and $\Lambda^{(m)}$ are chosen such that a prescribed number of variables are selected from each column of $\hat{V}^*$ and $\hat{V}^{(m)}$, $m=1,...,M$. 

\item Record the variables that are selected in $\hat{V}^*$ up to and including $d=d_0$ and in $\hat{V}^{(m)}$ up to and including $r=r_0$, $m=1,...,M$.

\item Repeat steps 1 to 3 $N$ times, where $N$ is at least 1000.

\item Compute the empirical selection probabilities for each variable in $\hat{V}^*$ and $\hat{V}^{(m)}$, $m=1,...,M$. Then rank the variables in each list according to the selection probabilities. 

\end{enumerate}

Note in step $2$, the number of variables to be selected in each column of $\hat{V}^*$ and $\hat{V}^{(m)}$, $m=1,...,M$, is a regularisation parameter. Nevertheless, \cite{meinshausen10} showed the variable rankings, especially the top ranking variables, were insensitive to the choice of these parameters which regularised the level of sparsity. In practice, the number of variables selected in each column of $\hat{V}^*$ and $\hat{V}^{(m)}$, $m=1,...,M$, is randomly picked and kept small. In the TwinsUK study, it is chosen to be $100$ since we would only be interested in the top few hundred probes which drove the shared and tissue-specific variability respectively.

\section{Illustration with simulated data} \label{sMVMF:simulation}

In this section we present simulation studies to characterise how the sMVMF method is able to distinguish between shared and tissue-specific variance. We simulate shared and tissue-specific variance patterns as illustrated by the middle and right panels in Figure \ref{sMVMF:fig:simuB_truevar}. We then test whether sMVMF correctly decomposes the total sample variance (left panel) whilst detecting variables contributing to the non-random variability within each variance component. We also compare sMVMF with two alternative methods: standard PCA and Levene's test \citep{gastwirth09} of the equality of variance between population groups. 

\subsection{Simulation setting}   \label{sMVMF:simulation:setting}

Our simulation study consists of 1000 independent experiments. In each experiment we simulate 3 data matrices or datasets (tissues) of dimension $n=100$ (samples) and $p=500$ (genes). Each simulated data matrix $X^{(m)}$ is obtained via: 
$$
X^{(m)} = Y^{(m)} + Z^{(m)} + E^{(m)},
$$ 
where $Y^{(m)}$ is a component designed to control the shared variance, $Z^{(m)}$ is introduced to control the tissue-specific variance, and $E^{(m)}$ is a random error. They are all $n \times p$ random matrices.  Since we ultimately wish to test whether our method is able to distinguish between signal and noise variables, we assume that only the first $30$ variables carry the signal, whereas the remaining $470$ only introduce noise. 

We suppose that the shared variability is controlled by the activation of $3$ latent factors, each regulating the variance of a different block of variables. To this end, we further group the $30$ signal variables into three blocks of $10$ normally distributed random variables each (numbered $1-10$,$ 11-20$, and $21-30$), as illustrated in Figure \ref{sMVMF:fig:simuB_heatmap1} (A). We design the simulations so that each of the first 30 variables in $Y$ has the same variance in different datasets; moreover, the variance decreases while moving from the first to the third block. Further details and simulation parameters are given in Appendix, Section \ref{appendix-simulation}. This procedure generates shared variance patterns that look like those reported in the middle panel of Figure \ref{sMVMF:fig:simuB_truevar}. 

The variables in $Z$ are also assumed to be normally distributed. They are generated such that exactly 10 of them have the largest variance across datasets. The resulting "mosaic" structure of the simulated variance patterns is illustrated in right panel of Figure \ref{sMVMF:fig:simuB_truevar}. The data matrices $Y^{(m)}$ and $Z^{(m)}$ are generated such that the total non-random sample variance of each variable in a tissue equals the sum of its shared and tissue-specific variances, which is also illustrated in Figure \ref{sMVMF:fig:simuB_truevar}. The random error term $E^{(m)}$ is generated from independent and identical normal distributions with zero mean and noise $\sigma_\epsilon^2$ for all variables in all datasets. We perform simulations on two settings: in setting I $\sigma_\epsilon^2=1$ and in setting II $\sigma_\epsilon^2=4$. As a result of this simulation design, we are able to characterise the true underlying architecture that explains the total sample variance.

\subsection{Simulation results}    \label{sMVMF:simulation:result}

The data generated in each experiment was analysed by fitting the sMVMF algorithm. To focus on the ability of the model to disentangle the true sources of variability, we take $d=3$ and $r=1$, which equal the true number of shared and tissue-specific LFs used to generate the data. The regularisation parameters $\Lambda^*$ and $\Lambda^{(m)}$ are tuned such that each PPJ consists of  $10$ variables, the true number of signal variables. 

For comparison, we propose two additional approaches that are able to identify variables featuring dataset-specific sample variances, although they do not attempt to model the shared variance. The first method consists of carrying out a separate PCA on each dataset; for each PCA/dataset, we then select the $10$ variables having the largest loadings in the first principal component. The second method consists of applying a standard Levene's test of equality of population variances independently for each variable, which is then followed by a Bonferroni adjustment to control the family-wise error rate; if a test rejects the null hypothesis at the $5\%$ significance level, we select the variable having the largest sample variance amongst the three datasets. 

\begin{figure}[!tpb]
 \centerline{\includegraphics[scale=0.3]{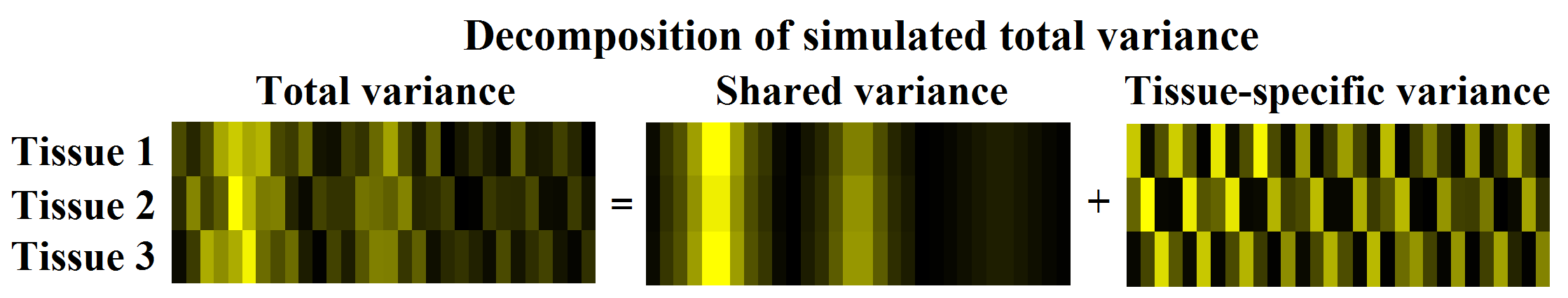}}
 \caption{Simulated patterns of sample variance: the total, non-random, sample variance of 30 signal-carrying random variables is generated so that it can be decomposed into the sum of shared and tissue-specific components. Rows correspond to tissues (datasets) and columns correspond to 30 variables. Brighter colours represent large variance and darker colours represent low variance. Although by construction the underlying shared and tissue-specific variances have very different patterns, sMVMF is able to discriminate between them.  }
 \label{sMVMF:fig:simuB_truevar} 
\end{figure}

By averaging across $1000$ experiments, we are able to estimate the probability that each one of the $30$ signal variables is selected by each one of the three competing methods. The heatmaps (A)-(C) in Figure \ref{sMVMF:fig:simuB_heatmap2} visually represent these selection probabilities for simulation setting I.  Here sMVMF perfectly identifies the variables that introduce dataset-specific variability. The results obtained using Levene's tests are somewhat similar, except for some variables in the first block (indexed $3-8$) and second block (indexed $14-17$). By reference to the middle panel of Figure \ref{sMVMF:fig:simuB_truevar},  it can be noted that these variables are precisely those featuring large shared variability by construction. On the other hand, the PCA-based approach performs poorly because it can only select variables that contribute to explaining the total sample variance, but is unable to capture dataset-specific patterns. This example is meant to illustrate the limitations of both univariate and multivariate approaches that do not explicitly account for factors driving shared and dataset-specific effects. sMVMF has been designed to address exactly these limitations. 

\begin{figure}[!tpb]
 \centerline{\includegraphics[scale=0.5]{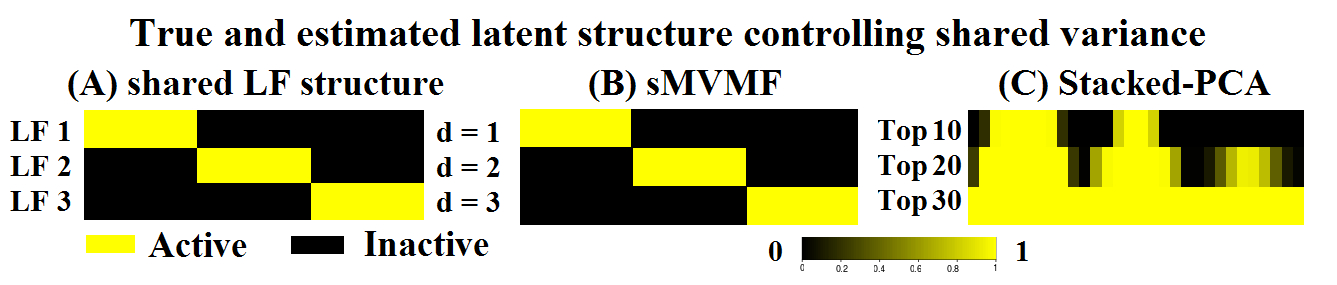}}
 \caption{Each latent factor (LF) is only active in a block of 10 signal- carrying variables, and controls the amount of variance of those variables that is shared amongst datasets. The (A) panel shows the true latent structure used to generate the data. Panels (B) and (C) show the estimated probabilities that each variable has been selected as signal-carrier using sMVMF and a stacked-PCA approach, respectively. sMVMF accurately captures the true shared LF structure whereas stacked-PCA tends to identify variables with large variance but fails to identify the LF structure.}
 \label{sMVMF:fig:simuB_heatmap1}
\end{figure} 

\begin{figure}[!tpb]
 \centerline{\includegraphics[scale=0.52]{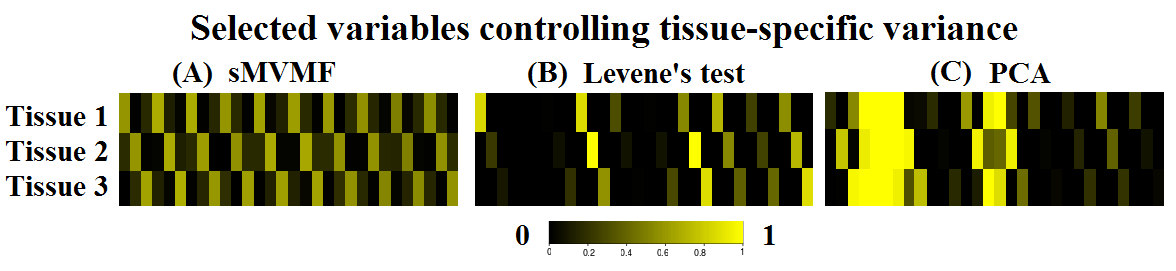}}
 \caption{Three different methods -- sMVMF, Levene's test and PCA --  are used to detect random variables whose variance pattern is dataset-specific.  Each heatmap represents the selection probabilities estimated by each method: (A) sMVMF produces patterns that closely match the true tissue-specific variances shown in the right panel of Figure \ref{sMVMF:fig:simuB_truevar}; (B) Levene's test performs well for variables those variance is mostly driven by tissue-specific factors, but fails to detect those variables having a strong shared-variance component; (C) The PCA-based method cannot distinguish between shared and tissue-specific variability, and fails to recover the true pattern.}
 \label{sMVMF:fig:simuB_heatmap2}
\end{figure} 



Both Levene's test and the individual-PCA approach are not designed to capture shared variance patterns. As a way of direct comparison with sMVMF we therefore propose an alternative PCA-based approach that has the potential to identify variables associated to the direction of largest variance across all three datasets. This method consists of performing a single PCA on a ``stacked'' matrix of dimension $(Mn) \times p$ containing measurements collected from all three datasets, and obtained by coalescing the rows of the three individual data matrices. By varying the cutoff value for thresholding the loadings of the first PC, we are able to select the top $10$, $20$, and $30$ variables. We shall refer to this approach as stacked-PCA.  

Results produced by sMVMF and stacked-PCA are summarised by the heatmaps (B) and (C) in Figure \ref{sMVMF:fig:simuB_heatmap1}, and can be directly compared to the true simulated patterns in (A). As expected, stacked-PCA tends to select variables having large total sample variances, whereas sMVMF can identify variables affected by each shared LF which jointly explain a large amount of variance. This example shows that sMVMF is able to identify the variables associated to the latent factors controlling the shared variance. 

We also carried out a simulation, based upon the same setting, with smaller signal-to-noise ratio, i.e. by sampling the random error terms in $E^{(m)}$ from independent normal distributions having larger variance. The results were very similar to the previous setting, except that Levene's test was hardly able to identify any tissue-specific genes. The heatmaps summarising model performances are given in Appendix, Section \ref{appendix-simulation2}.

\section{Application to the TwinsUK cohort} \label{sMVMF:application}

\subsection{Data preparation}

TwinsUK is one of the most deeply phenotyped and well-characterised adult twin cohort in the world \citep{moayyeri13b}. It has been widely used in studying the genetic basis of aging procession as well as complex diseases \citep{codd13}. More importantly, it contains a broad range of `omics' data including genomic, epigenomic and transcriptomic profiles amongst others \citep{bell12}. In this study, we focus on comparing the variance of mRNA expressions in adipose (subcutaneous fat), lymphoblastoid cell lines (LCL), and skin tissues. The microarray data used in this study were obtained from the Multiple Tissue Human Expression Resource \citep{nica11}, with participants being recruited from the TwinsUK registry. Peripheral blood samples were artificially transformed from mature blood cells by infecting them with the Epstein-Barr virus \citep{glass13}. All tissue samples were collected from $856$ female Caucasian twins ($154$ monozygotic twin pairs, $232$ dizygotic twin pairs and $84$ singletons) aged between $39$ and $85$ years old (mean $62$ years). Genome-wide expression profiling was performed using Illumina Human HT-12 V3 BeadChips, which included $48,804$ probes. Log2-transformed expression signals were normalized per tissue using quantile normalization of the replicates of each individual followed by quantile normalization across all individuals, as described in \cite{nica11}. In addition, we also had access to 450K methylation data of the same adipose biopsies profiled using Infinium HumanMethylation 450K BeadChip Kit \citep{wolber14}. We only retained probes whose expression levels were measured in all three tissues, and removed subjects comprising unmeasured expressions in any tissue. Using the same notation introduced before, this resulted in three data matrices each of dimension $n=618$ and $p=26017$. For each probe in each tissue, a linear regression model was fitted to regress out the effects of age and experimental batch, following the same procedure as in \cite{grundberg12}. Residuals in adipose, LCL, and skin tissues were arranged in $n \times p$ matrices $X^{(1)}$, $X^{(2)}$, $X^{(3)}$, respectively, for further analysis using the proposed multiple-view matrix factorisation method.

\subsection{Experimental results}

Non-sparse MVMF was initially fitted for all combination of parameter pairs $(d,r)$ in a grid. 
For each model fit, we computed the percentage of variance explained in each tissue. These are shown in the 3D bar charts presented in Appendix, Section \ref{appendix-plot}, Figure \ref{fig:fatHeatmap}. The percentages of variance explained varied between 25.2\% ($d=r=1$, LCL) and 87.3\% ($d=r=160$, skin). The following analyses are based on the $d=r=3$ setting, which explains at least 40\% of expression variance across tissues. Given that there are more than 26000 probes, and this is much larger than the sample size, this choice of parameters offers a good balance between dimensionality reduction and retaining a large portion of total variance. Although two other combinations of $(d,r)$, i.e. $(2,4)$ and $(4,2)$, also explain a similar amount of total variance, we have found that the gene ranking results are not extremely sensitive to these values. For more details on this sensitivity analysis, see Appendix, Section \ref{appendix-dr}.

\begin{figure}[!h]
 \centerline{\includegraphics[scale=0.4]{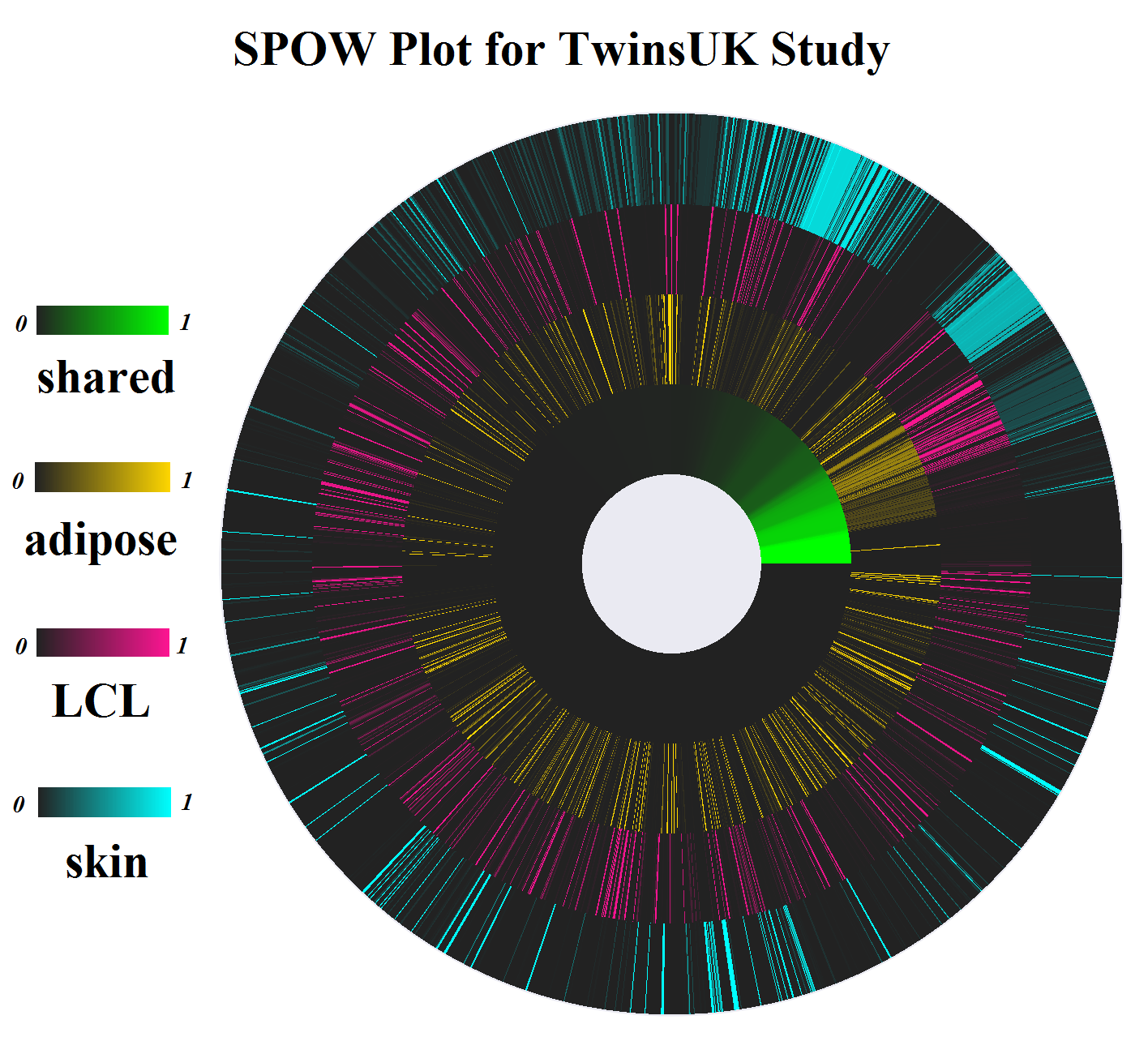}}
 \caption{TwinsUK study: resulting SPOW plot. The wheel comprises four rings, which correspond to shared, adipose-, LCL-, and skin-specific variability from the inner ring. It is also evenly divided into 3274 fan slices, corresponding to 3274 mRNA expression probes that are selected at least once in all subsamples. Probes are re-ordered by their selection probabilities in the transformation matrix in the shared component. Brighter colour denotes higher probability, whereas darker colour denotes lower probability. We are particularly interested in probes with high selection probability exclusively in one ring.}
 \label{sMVMF:fig:application_selection}
\end{figure} 


The sparse version of our model, sMVMF, to each subsample in stability selection procedure to rank gene expressions explaining a large amount of shared and tissue-specific variances respectively. A detailed description of the procedure is presented in Section \ref{sMVMF:tuning}. In summary, $1000$ random subsamples were generated each consisting of $309$ subjects randomly and independently sampled without replacement from a total of $618$. No twin pair was included in any subsample in order to remove possible correlations due to zygosity. sMVMF was fitted to each subsample, where the sparsity parameters were fixed such that each column of the transformation matrices comprised exactly 100 non-zero entries. There were $3274$ mRNA expression probes that were selected at least once from any of the transformation matrices. 

Probes that explain a large amount of expression variance exclusively in one tissue are of particular interest. To make such probes visually discernible we propose a new visualisation tool, the SPOW (Selection PrObability Wheel) plot. The plot in Figure \ref{sMVMF:fig:application_selection} consists of $3274$ fan slices corresponding to probes that are selected at least once in all subsamples, re-ordered by their selection probabilities in $\hat{V}^*$. The wheel is further divided into four rings, representing shared, adipose-, LCL-, and skin tissue, respectively. Each ring is assigned a unique colour spectrum to illustrate selection probabilities of the probes: brighter colours denote a higher probability and darker colours denote a lower probability. Probes featuring exclusively shared or tissue-specific variability can be found along the radii where only one part is painted in a bright colour and the other three parts are colored in black. The SPOW plot{\color{black}s} for the top 200 probes that explain shared and tissue-specific variability respectively are presented in Appendix, Section \ref{appendix-plot}, Figures \ref{sharedSPOW} to \ref{skinSPOW}, where such probes can be more easily captured. 

Four groups of mRNA expressions were selected for further investigation, corresponding to shared-exclusive, adipose-, LCL-, and skin-exclusive expressions. Each group consisted of probes whose selection probabilities were larger than 0.5 in the corresponding transformation matrix and less than 0.005 in the other transformation matrices. These thresholds were set to give a manageable number of featured gene probes while tolerating occasional selection in the other groups.  This procedure selected $294$ genes for further study, including 114 adipose-exclusive, 83 LCL-exclusive, 64 skin-exclusive, and 33 shared-exclusive genes. We summarise the results in Table \ref{sMVMF:tab:mutherSummary}. A Venn-diagram representation of the results is given in Appendix, Section \ref{appendix-venn}.

\begin{table}[!h]  
\centering
\caption{TwinsUK study: summary of results. There are additionally 33 shared-exclusive genes.}    \label{sMVMF:tab:mutherSummary}
{\begin{tabular}{lclclclclc}
\hline
& \% of variance    & \% of variance   & Number  & Number \\
&  explained by   &  explained by    & of tissue-  & of tissue- \\
&  tissue-specific   & shared    &  exclusive   &  exclusive   \\
&  component        & component    &  probes    &  genes   \\
\hline
Adipose   & 27.0 & 14.7  & 132 & 114 \\
LCL          & 30.8 & 12.1 & 91 & 83   \\
Skin         & 32.6 & 11.5 & 74 & 64   \\
\hline
\end{tabular}}
\end{table}

For each tissue, we performed an enrichment test by overlapping genes in our list with genes contained in the TiGER and VeryGene databases to examine the extent of agreement. In addition, a Gene Ontology (GO) biological process pathway enrichment test \citep{ashburner00} and a Cytoscape pathway (CP) analysis \citep{saito12}
were carried out to reveal the function of the pathways which the 261 tissue-exclusive genes belonged to, and FDR-corrected p-values were reported (See Supplementary Material, Table T1 and T2 for full results). Below we present test results for each group of genes separately for each tissue. We also report the selection probability (SP) for some selected probes.

\subsubsection*{Skin-exclusive genes.}  15 of the 64 genes from our skin-exclusive list are contained in the combined TiGER/VeryGene list, giving rise to significant enrichment of our list with Fisher exact test p-value $p<10^{-16}$. The overlapping genes include serine protease family genes KLK5 (SP:  $1.000$) and KLK7 (SP: $1.000$), which are highly expressed in the epidermis and related to various skin conditions, such as cell shedding (desquamation) \citep{brattsand99}. Another member ALOX12B (SP: $1.000$) controls producing 12R-LOX, which adds an oxygen molecule to a fatty acid to produce the 12R-hydroperoxyeicosatetraenoic acid that has major function in the skin cell proliferation and differentiation \citep{juanes09}. The skin-exclusive genes have also been found significantly enriched in two biological processes, namely epidermis development and cell-cell adhesion ($p<0.001$ and $p=0.03$, respectively).

\subsubsection*{LCL-exclusive genes.} LCLs are not natural human cells: they are laboratory induced immortal cells that have abnormal telomerase activity and tumorigenic property \citep{sie09}. Since neither TiGER nor VeryGene assessed transcriptomic profile in LCL cells, we obtained LCLs data from \cite{li10}, in which the authors compared LCLs expression profile in four human populations and reported 282 LCL specific expression genes. $9$ of those genes are contained in our LCL-exclusive gene list, giving a Fisher exact test $p<10^{-16}$. These include CDK5R1 (SP: $0.961$) and HEY1 (SP: $1.000$), which are key genes in the transformation of B lymphocytes to LCLs \citep{zhao06}. Pathway analysis of the LCL-exclusive genes reveals several aging and cell-death related pathways such as regulation of telomerase (CP enrichment test, $p=0.014$), small cell lung cancer (CP enrichment test, $p=0.019$), and cell cycle checkpoints (CP enrichment test, $p=0.021$). These results show that our tissue-exclusive genes represent tissue unique molecular functions and biological pathways, which may be used to validate known pathways or discover new biological mechanisms.    

\subsubsection*{Adipose-exclusive genes.} ApoB (SP: $1.000$) is the only member in our adipose-exclusive list which is also contained in the list of known adipose-specific expression genes (Fisher exact test, $p=0.05$). ApoB is one of the primary apolipoproteins that transport cholesterol to peripheral tissues \citep{knott86} and it has been widely linked to fat formation \citep{riches99}. In adipose, the selected genes are found significantly enriched in triglyceride catabolic process pathway ($p=0.022$), which is in line with the fact that adipose tissue is the major storage site for fat in the form of triglycerides. Pathway analysis reveals that genes in the adipose-exclusive list are significantly enriched in triglyceride catabolic process pathway ($p=0.022$), which agrees with the fact that adipose tissue is the major storage site for fat in the form of triglycerides. In addition, these genes are enriched in inflammation pathways, such as lymphocyte chemotaxis ($p=0.016$) and neutrophil chemotaxis ($p=0.027$). This coincides with previous findings of the complex and strong link between metabolism and immune system in adipose tissue \citep{tilg06}.

For this tissue we were also able to further investigate the causes for the observed adipose-exclusive gene expression variability. One possible explanation could be that environmental factors influenced an individual's epigenetic status, which subsequently regulated gene expression \citep{razin91}. As a mediator of gene regulatory mechanisms, DNA methylation is crucial to genomic functions such as transcription, chromosomal stability, imprinting, and X-chromosome inactivation \citep{lokk14}, 
which consequently influence an individual's tissue development \citep{ziller13}. 
It thus seemed reasonable to hypothesise that the expression of tissue-exclusive genes could be modified by their methylation status in the same tissue. 

We sought to identify genes featuring a statistically significant linear relationship between the gene's methylation profile and its expression value from the same tissue. In adipose biopsies, where both transcriptome and methylation data is available, we found that $68.4\%$ ($78$ out of $114$ genes) of the genes had expression levels significantly associated with their methylation status using a linear fit (Bonferroni correction, $p<0.05$) (See Supplementary Material, Table T3, for full lists). We then wanted to assess whether a similar number of linear associations could be found by chance only by randomly selecting any genes, not only those that feature adipose-exclusive variability, and testing for association between gene expression and methylation levels. This was done by randomly extracting the same, fixed number ($132$) of expression probes and corresponding methylation levels from adipose tissue, and fitting a linear model as before. By repeating this experiment $1000$ times, we obtained the empirical distribution reported in Appendix, Section \ref{appendix-plot}, Figure \ref{fig:fatMethyMean}. This distribution suggested that all the proportions were below $0.2$, compared to our observed proportion of $0.684$, which provided overwhelming evidence that DNA methylation was an important factor affecting the expression of the tissue-exclusive genes. It was notable that the adipose-exclusive variability of ApoB was regulated by methylation at 50bp upstream of the Transcriptional Starting Site (linear fit, $p=2.1\times 10^{-5}$), which agreed with the findings that the promoter of ApoB has tissue-specific and species-specific methylation property \citep{apostel02}. Apart from ApoB, we also found that methylation in Syk was associated with Syk expression level, which was potentially involved in B cell development and cell apoptosis \citep{ma10}.

\section{Conclusion and Discussion}    \label{sMVMF:discussion}

The proposed sMVMF method facilitates the comparison of gene expression variances across multiple tissues. The primary challenge of this task arises from the interference between substantial co-variability of gene expressions across all tissues and substantial variability of gene expressions featured only in specific tissues. Characterising tissue-specific variability can shed light on the biological processes involved with tissue differentiation. Analysing shared variability can potentially reveal genes that are involved in complex or basic biological processes, and may as well enhance the estimation of tissue-specific variability. 

sMVMF has been used here to compare gene expression variances in three human tissues from the TwinsUK cohort. 261 genes having substantial expression variability exclusively featured in one tissue have been identified. Enrichment tests showed significant overlaps between our lists of tissue-exclusive genes and those reported in the TiGER and VeryGene databases, which were established by comparing mean expression levels. This confirms the link between tissue-specific expression variance and the biological functions associated with particular tissues. In future work, it would be interesting to explore the functions of the tissue-exclusive genes from our list that have not been reported in existing databases. We further showed adipose-exclusive expression variability was driven by an epigenetic effect. Using these results as a guiding principle, we expect our methods and results could improve efficiencies in mapping functional genes by reducing the multiple testing and enhancing the knowledge of gene function in tissue development and disease phenotypes. Future works would consist of investigating the outcome of tissue-exclusive expression variability, for which we can perform association studies between expressions of tissue-exclusive genes and disease phenotypes related to adipose and skin tissues.

\section*{Funding}
The Biological Research Council has supported ZW (DCIM-P31665) and the TwinsUK study. We also thank the European Community's Seventh Framework Programme (FP7/2007-2013) and the National Institute for Health Research (NIHR) for their support in the TwinsUK study.

\bibliographystyle{natbib}
\bibliography{sMVMF_bioinf_ref}

\begin{appendices}

\section{Simulation setting}   \label{appendix-simulation}

As introduced in Section \ref{sMVMF:simulation:setting} of the main text, variance of the first 30 variables (columns) in the random matrices $Y^{(m)}$ ($m=1,2,3$) are controlled by three latent factors: $H_1$, $H_2$, $H_3$, which are real valued univariate random variables generated from independent normal distributions as follows:
\begin{eqnarray}    \label{sMVMF:H-distribution}
H_1 \sim \text{$\mathcal{N}$}(0,5^2) ~~; ~~ H_2 \sim \text{$\mathcal{N}$}(0,3.5^2) ~~; ~~ H_3 \sim \text{$\mathcal{N}$}(0,2^2)     
\end{eqnarray}
where $\mathcal{N}(\mu,\sigma^2)$ refers to normal distribution with mean $\mu$ and standard deviation $\sigma$. 

Variance of the first 30 variables (columns) in the random matrices $Z^{(m)}$ ($m=1,2,3$) is controlled by three latent factors: $h_1$, $h_2$, $h_3$, where $h_m$ only affects $Z^{(m)}$. These latent factors are also generated from independent normal distributions:
\begin{eqnarray}    \label{sMVMF:h-distribution}
h_1 \sim \text{$\mathcal{N}$}(0,2.8^2) ~~; ~~ h_2 \sim \text{$\mathcal{N}$}(0,3.2^2) ~~; ~~ h_3 \sim \text{$\mathcal{N}$}(0,3^2)    
\end{eqnarray}
The latent variables in \eqref{sMVMF:H-distribution} and \eqref{sMVMF:h-distribution} control the variance of the first 30 variables in $Y$ and $Z$ via some constant factors which we shall define. Specifically, each value in the first 30 columns of $Y$ is obtained by multiplying one latent variable from $\{H_1, H_2, H_3\}$ with a constant factor from one of the two row vectors $\alpha$ or $\beta$, so that the variance pattern in $Y^{(m)}$ is precisely as is illustrated in the middle panel in Figure 2 of the main paper. Similarly, each value in the first 30 columns of $Z$ is obtained by multiplying one latent variable from $\{h_1, h_2, h_3\}$ with a constant factor from one of the row vectors $\gamma_1$, $\gamma_2$, $\gamma_3$, such that the variance pattern in $Z^{(m)}$ is precisely as is illustrated in the right panel in Figure 2 of the main paper. The details are given as follows:
\begin{equation}
\begin{gathered}
\alpha = (0.3,0.5,0.6,0.8,1,1,0.8,0.6,0.5,0.3)   \\ 
\beta = (0.6,0.7,0.8,0.9,1,1,0.9,0.8,0.7,0.6)   \\
\gamma_1 = (v_1,v_1,v_1,v_1,v_1), ~~ \text{where} ~~ v_1=(1,1/3,2/3,1,2/3,1/3)    \\
\gamma_2 = (v_2,v_2,v_2,v_2,v_2), ~~ \text{where} ~~ v_2=(2/3,1,1/3,1/3,1,2/3)    \\
\gamma_3 = (v_3,v_3,v_3,v_3,v_3), ~~ \text{where} ~~ v_3=(1/3,2/3,1,2/3,1/3,1)    
\end{gathered}
\end{equation}
Let $Y^{(m)}_{i,j}$ denote the $(i,j)^{th}$ entry of $Y^{(m)}$. Our simulated data are generated as follows: for $i=1,2,...,100$ and for $m=1,2,3$:
\begin{enumerate}
\item Generate $H_1$, $H_2$, $H_3$, $h_1$, $h_2$, and $h_3$ according to \eqref{sMVMF:H-distribution} and \eqref{sMVMF:h-distribution}.
\item Generate $E^{(m)}_{i,1:500}$ from independent normal distributions with zero mean and variance $\sigma_\epsilon^2$, where $\sigma_\epsilon^2=1$ in setting I and $\sigma_\epsilon^2=4$ in setting II.
\item Compute/Set: 
$$Y^{(m)}_{i,1:10}= \alpha \cdot H_1$$
$$Y^{(m)}_{i,11:20}= \alpha \cdot H_2$$
$$Y^{(m)}_{i,21:30} = \beta \cdot H_3$$
$$Y^{(m)}_{i,31:500} = 0$$
$$Z^{(m)}_{i,1:30} = \gamma_m \cdot h_m$$
$$Z^{(m)}_{i,31:500} = 0$$
\end{enumerate}
Finally, compute: $X^{(m)} = Y^{(m)} + Z^{(m)} + E^{(m)}$.

\section{Additional simulation}    \label{appendix-simulation2}

\begin{figure}[H]
 \centerline{\includegraphics[scale=0.5]{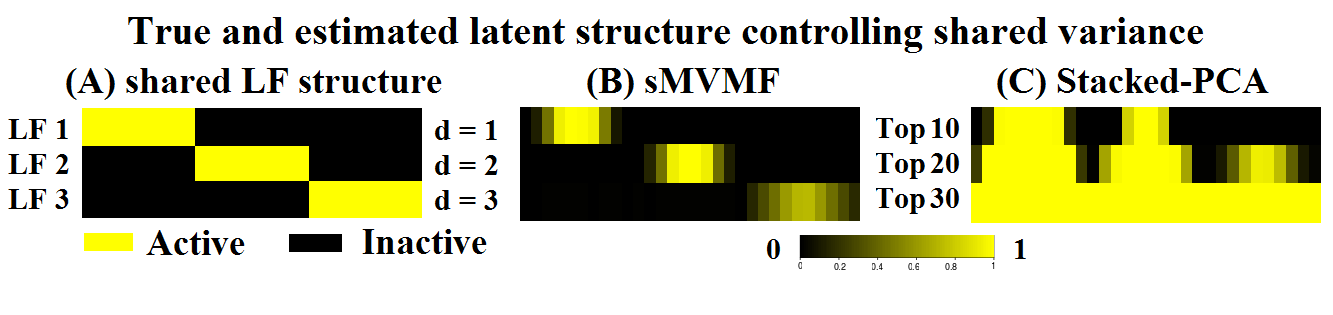}}
 \caption{Each latent factor (LF) is only active in a block of 10 signal-carrying variables, and controls the amount of variance of those variables that is shared amongst datasets. The (A) panel shows the true latent structure used to generate the data. Panels (B) and (C) show the estimated probabilities that each variable has been selected as signal-carrier using sMVMF and a stacked-PCA approach, respectively. sMVMF best captures the true shared LF structure whereas stacked-PCA tends to identify variables with large variance but fails to identify the LF structure.}
 \label{fig:simuC_heatmap1}
\end{figure}

\begin{figure}[H]
 \centerline{\includegraphics[scale=0.52]{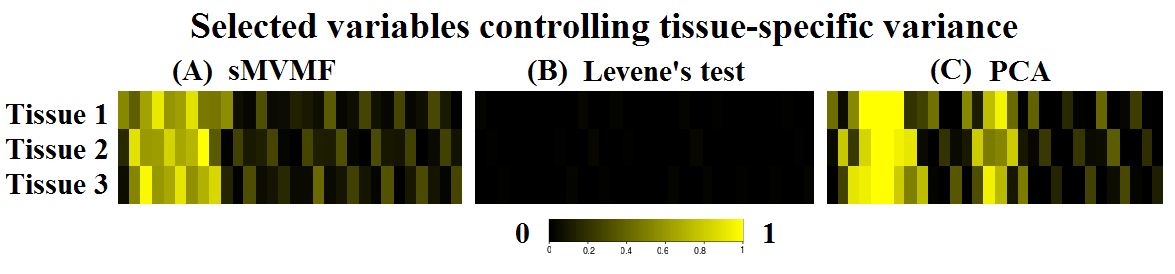}}
 \caption{Three different methods -- sMVMF, Levene's test and PCA --  are used to detect random variables whose variance pattern is dataset-specific.  Each heatmap represents the selection probabilities estimated by each method: (A) sMVMF produces patterns that best match the true tissue-specific variances shown in the right panel of Figure $1$ in the main paper; (B) Levene's test hardly detects any gene whose variance is significantly larger than the corresponding genes in the other tissues due to increased noise level; (C) The PCA-based method cannot distinguish between shared and tissue-specific variability, and fails to recover the true pattern on variables with large shared variance.}
 \label{fig:simuC_heatmap2}
\end{figure}

In this additional simulation we use the same settings as in the previous section except that $E^{(m)}$ is generated from independent normal distributions with zero mean and variance $4$. The same type of heatmaps as in Figure $2$ and $3$ of the main paper are produced and presented in Figure \ref{fig:simuC_heatmap1} and \ref{fig:simuC_heatmap2} respectively. We can visually conclude that sMVMF remains the best model in identifying the variables which drive shared and tissue-specific variance. Remarkably, Levene's test hardly detects any genes whose variance is significantly larger than the corresponding genes in the other tissues due to increased noise level.

\section{Robustness study on the choice of $(d,r)$}    \label{appendix-dr}

To investigate the robustness of $(d,r)$ on selected (shared- and tissue- exclusive) genes would require re-running the full analysis on all $(d,r)$ pairs on the $11\times11$ grid considered in our analysis, which would involve very intensive computation. Here we present a study in smaller scale in which we restrict the total amount of variance explained in adipose tissue to about $42\%$, and this gives us three pairs of $(d,r)$: $(3,3)$ which was the pair used to fit the sMVMF to identify shared- and tissue- exclusive genes in the paper, $(2,4)$ and $(4,2)$. We present the percentages of shared and tissue-specific variance explained for these three combinations of $(d,r)$ in Table \ref{tab:drsummary}. The figures in LCL and skin tissues are very similar ($\pm 3\%$) to the adipose tissue for each $(d,r)$

\begin{table}[H]  
\caption{Percentage of variance explained in adipose tissue}    \label{tab:drsummary}
\center
{\begin{tabular}{lclclclc}
\hline
 $(d,r)$   &  By shared component     & By tissue-specific component    &  Total \\
\hline
 $(3,3)$   & 14.7 & 27.0 & 41.7  \\
 $(2,4)$   & 10.3 & 32.6 & 42.9    \\
 $(4,2)$   & 23.3 & 18.4 & 41.7    \\
\hline
\end{tabular}} 
\end{table}

Notably, although the percentages of explained variance are approximately equal for the three combinations considered, the percentages within each component (shared and tissue-specific variances) vary substantially, in particular between $(2,4)$ and $(4,2)$. Therefore, this incomplete comparison seems to give a valid illustration of the robustness of gene selection results on the full grid of $(d,r)$.

\begin{figure}[!h]
 \centerline{\includegraphics[scale=0.16]{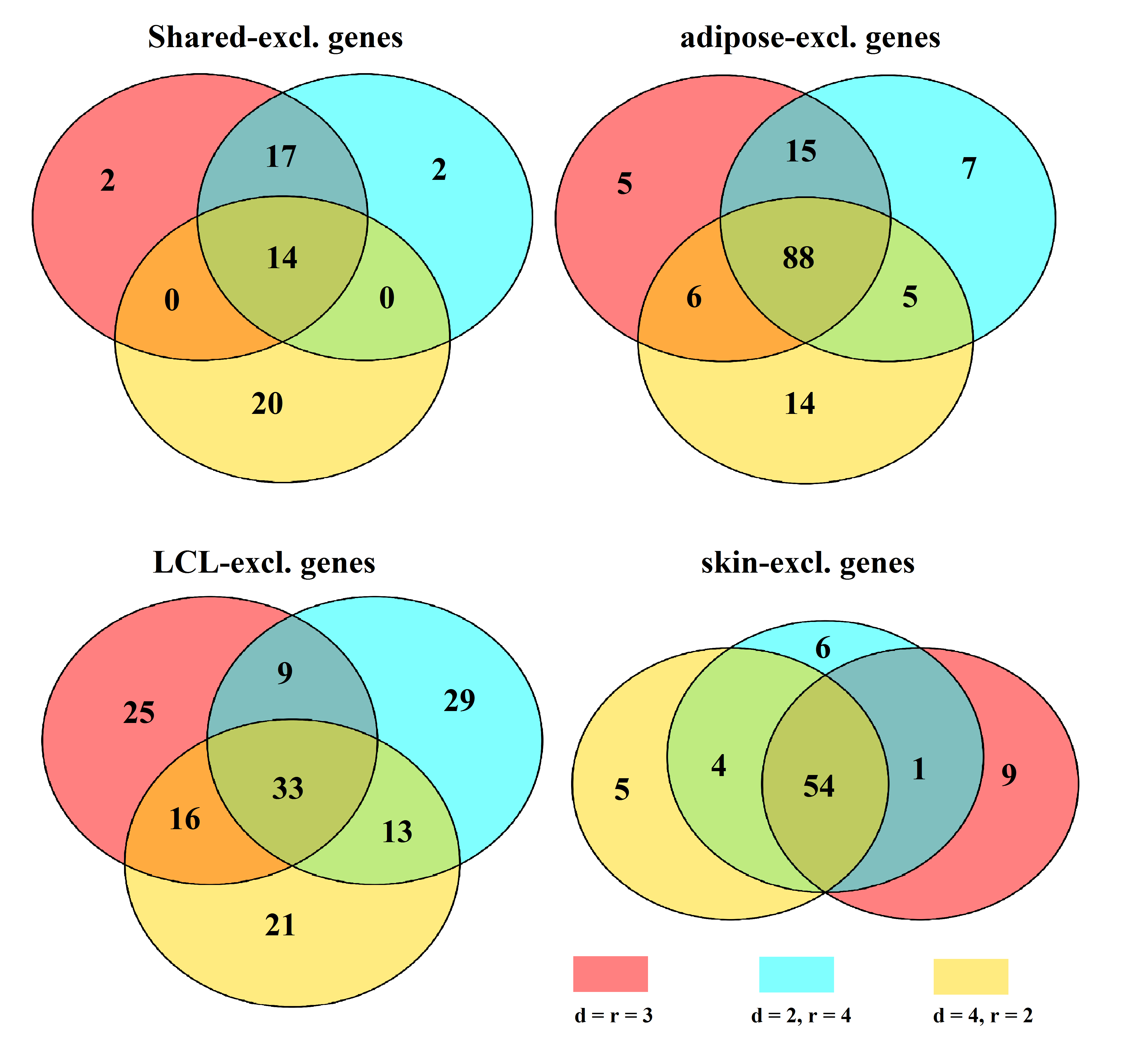}}
 \caption{The Venn diagrams summarise the overlaps of the shared- and tissue- exclusive genes selected for three combinations of $(d,r)$ pairs which explain about $42\%$ of the total variance in the adipose tissue. These plots showed that adipose- and skin- exclusive genes were very robust to the choice of $(d,r)$ and shared- and LCL- exclusive genes were fairly robust.}
\label{reply:venn}
\end{figure}

To evaluate the robustness of the shared- and tissue- exclusive genes with respect to the choice of $(d,r)$, we repeated our analysis for $(d,r)=(2,4)$ and $(d,r)=(4,2)$ in the same way as for $(d,r)=(3,3)$. We adjusted the selection criteria (threshold of selection probabilities) following the same principle as introduced in the paper so that the same number of shared- and tissue- exclusive genes ($\pm 1$ when there were ties) were selected as in the lists for $(d,r)=(3,3)$. We present the Venn diagrams summarising the overlaps between the three combinations of $(d,r)$ parameters in Figure \ref{reply:venn}. 

The results showed that adipose- and skin- exclusive genes were very robust to the choice of $(d,r)$ since more than $77\%$ of genes appeared in the lists obtained from all three combinations of $(d,r)$. The shared-exclusive genes were fairly robust to the choice of $(d,r)$ in that there were more than $90\%$ of overlaps between the lists obtained from $(d,r)=(3,3)$ and $(d,r)=(2,4)$, and about $40\%$ of overlaps with the list obtained from $(d,r)=(4,2)$. However, the percentage of overlaps would increase to $70\%$ if we restrain the comparison to the top $20$ shared-exclusive genes. For LCL-exclusive genes, there were about $40\%$ of overlaps among the three pairs of $(d,r)$. Moreover, the highlighted genes mentioned in the main paper were all retained in the lists of genes selected using the other combinations of $(d,r)$, except for the LCL-exclusive gene CDK5R1 which was absent from $(d,r)=(4,2)$. We therefore conclude that given the substantial difference in the percentages of shared and tissue-specific variance explained using different combinations of $(d,r)$, the lists of shared- and tissue- exclusive genes were robust to the choice of $(d,r)$, in particular if such lists were small.

\section{Venn-diagram analysis}    \label{appendix-venn}

\begin{figure}[H]
 \centerline{\includegraphics[scale=0.4]{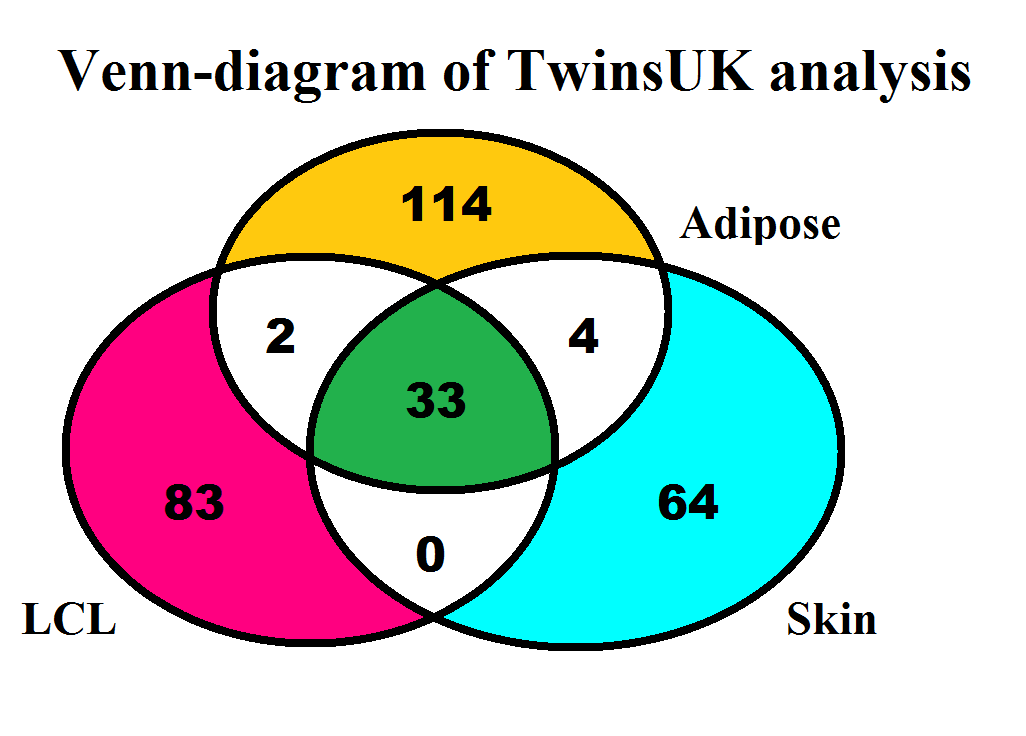}}
 \caption{The Venn diagram shows that there were 114 adipose-exclusive, 83 LCL-exclusive, 64 skin-exclusive, and 33 shared-exclusive genes extracted from our analysis. Using the SPOW plot in Figure 4 of the main paper, we were also able to identify 2 genes (``AQP9'' and ``TYMP'') which drove tissue-specific variability in adipose and LCL tissues but not in skin; in addition we also identified 4 genes (``CCND1'', ``GPC4'', ``GSDMB'', and ``TUBB2B'') which drove tissue-specific variability in adipose and skin tissues but not in LCL.}
 \label{fig:venn}
\end{figure}

We present a Venn diagram  in Figure \ref{fig:venn} summarising our findings from the TwinsUK analysis. As mentioned in the main paper, we identified 114 adipose-exclusive, 83 LCL-exclusive, and 64 skin-exclusive genes. In addition, 33 genes which drove the shared variability across all three tissues yet without driving tissue-specific variability in any tissue were identified. Moreover, 2 genes (``AQP9'' and ``TYMP'') were identified to have driven adipose- and LCL-specific variance but not skin-specific variance, while 4 genes  (``CCND1'', ``GPC4'', ``GSDMB'', and ``TUBB2B'') were found to have driven adipose- and skin-specific variance but not LCL-specific variance. 

\section{Plots}    \label{appendix-plot}
\begin{figure}[h]
 \centerline{\includegraphics[scale=0.24]{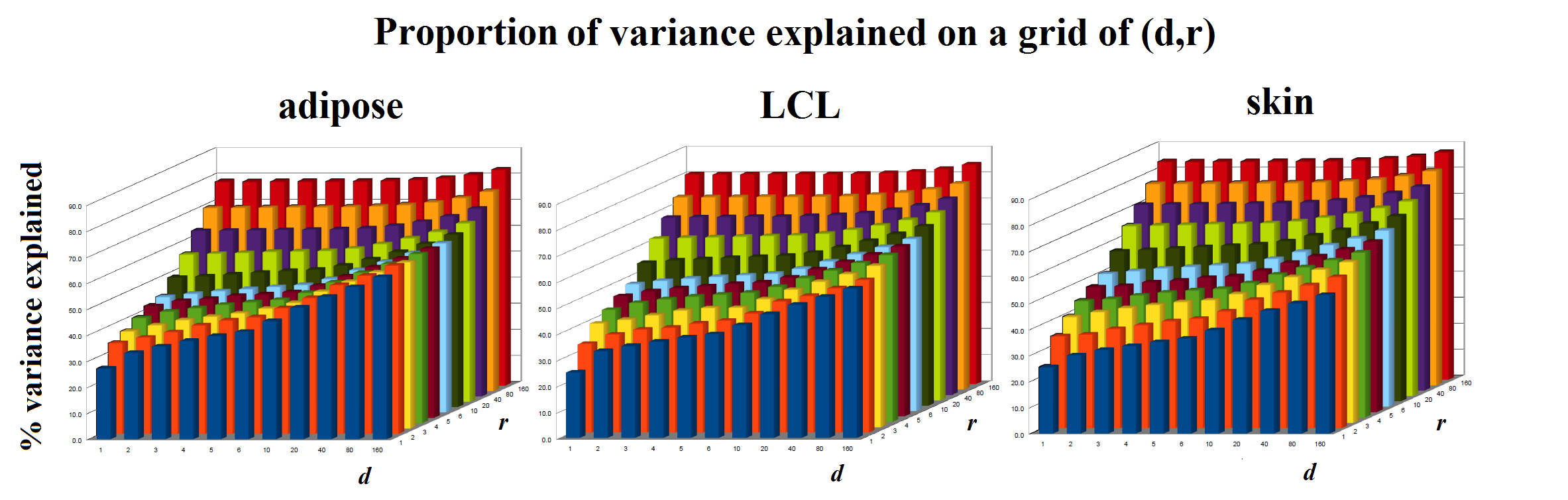}}
 \caption{TwinsUK study: 3D boxplot showing the percentage of expression variance explained in adipose, LCL, and skin tissues on a grid of $(d,r)$ using the non-sparse MVMF. $d$ is the total number of PPJs in the shared variance component, and $r$ is the total number of PPJs in the tissue-specific variance component. The percentages vary between $25.2\%$ ($d=r=1$, LCL) and $87.3\%$ ($d=r=160$, skin).}
 \label{fig:fatHeatmap}
\end{figure} 

\begin{figure}[!tpb]    
 \centerline{\includegraphics[scale=0.4]{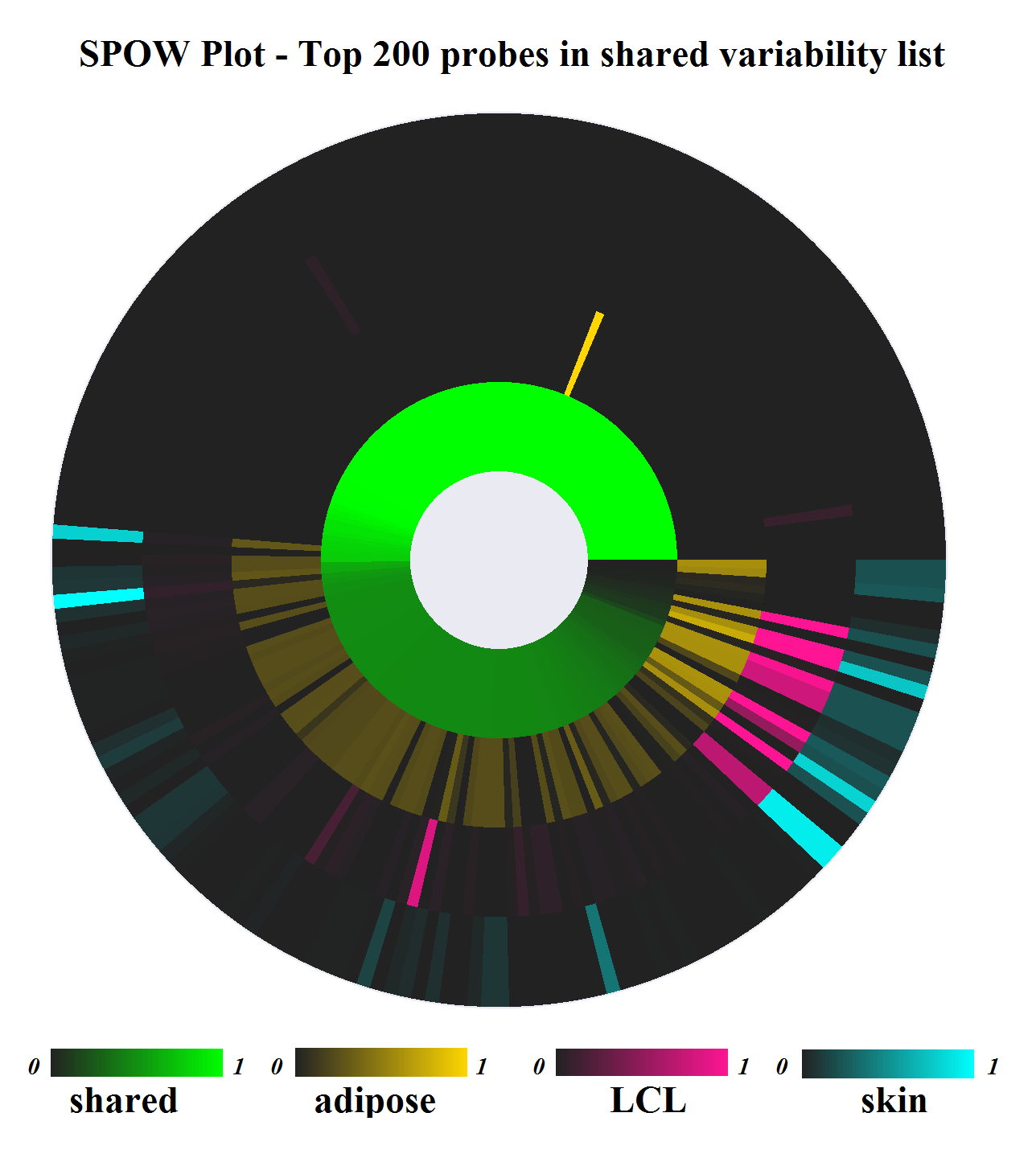}}
 \caption{TwinsUK study: SPOW plot $(d=r=3)$. The wheel contains the top 200 most frequently selected probes from the transformation matrix in the shared component. We extract probes with bright colour in the shared variability (green) ring and dark colours in the other rings.}
\label{sharedSPOW}

\end{figure} 

\begin{figure}[!tpb]
 \centerline{\includegraphics[scale=0.4]{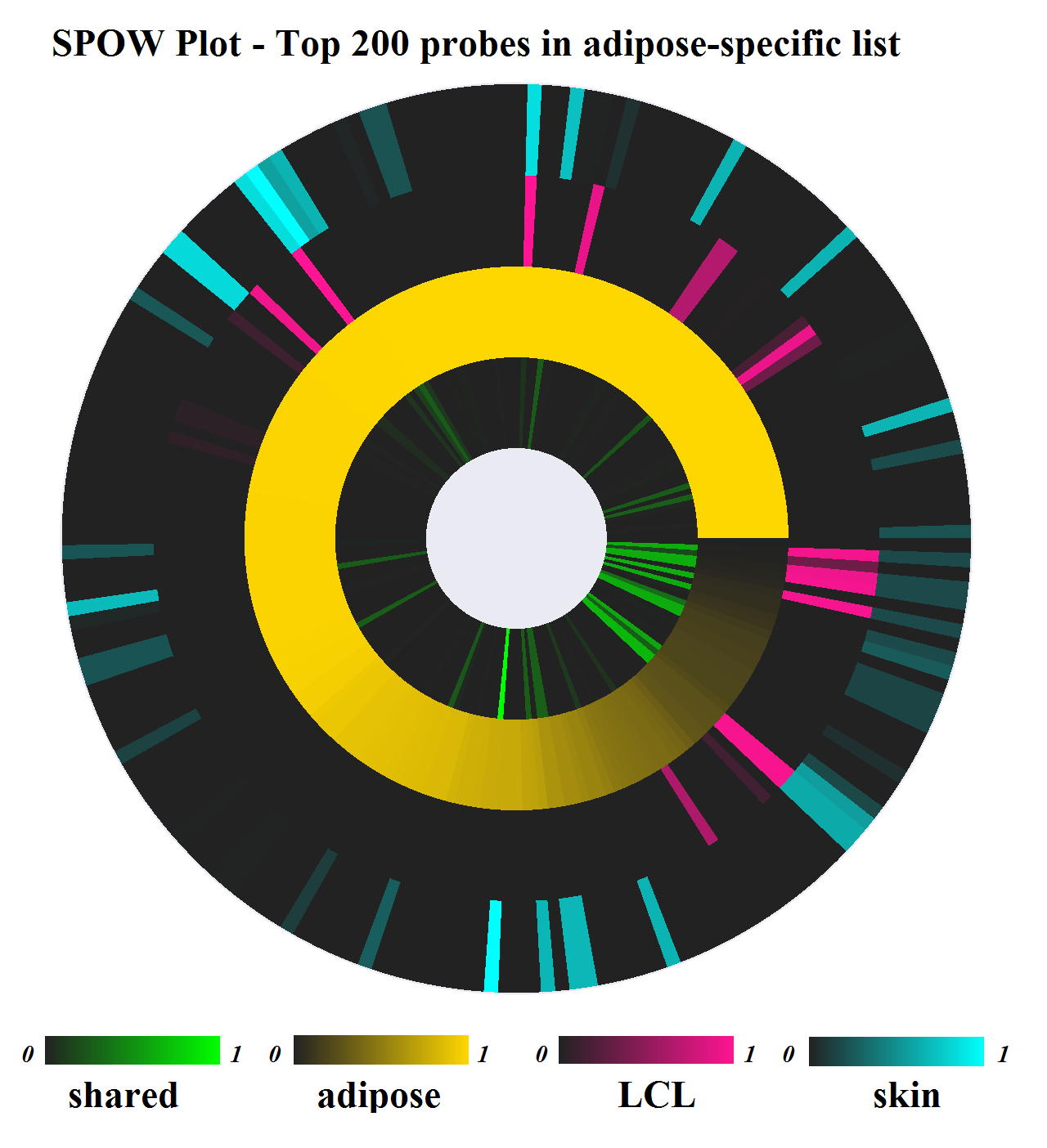}}
 \caption{TwinsUK study: SPOW plot $(d=r=3)$. The wheel contains the top 200 most frequently selected probes from the transformation matrix in the adipose-specific component using sMVMF. We extract probes with bright colour in the adipose-specific (yellow) ring and dark colours in the other rings.}
\label{fatSPOW}
\end{figure} 

\begin{figure}[!tpb]
 \centerline{\includegraphics[scale=0.4]{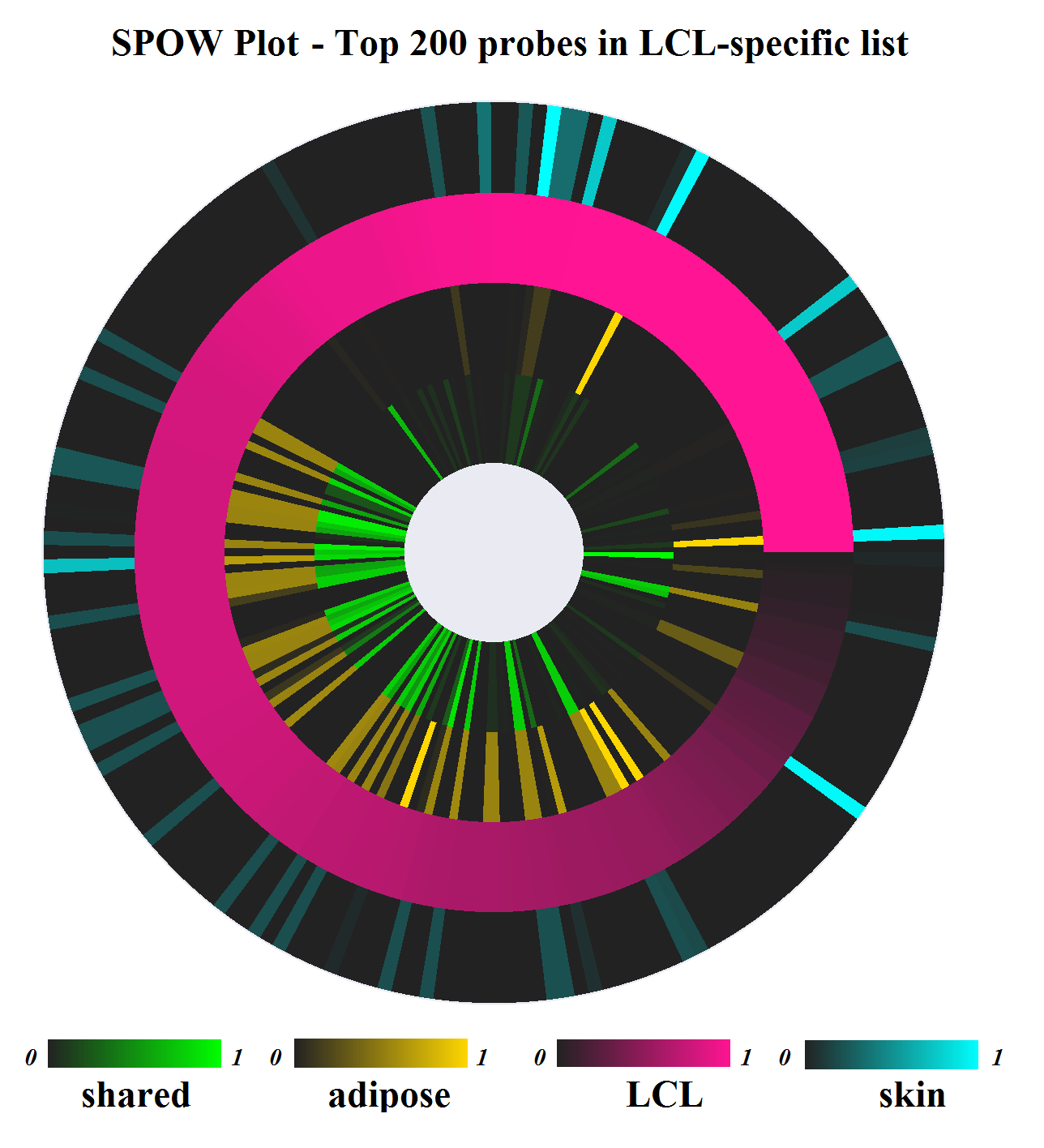}}
 \caption{TwinsUK study: SPOW plot $(d=r=3)$. The wheel contains the top 200 most frequently selected probes from the transformation matrix in the LCL-specific component using sMVMF. We extract probes with bright colour in the LCL-specific (purple) ring and dark colours in the other rings.}
\label{lclSPOW}
\end{figure} 

\begin{figure}[!tpb]
 \centerline{\includegraphics[scale=0.4]{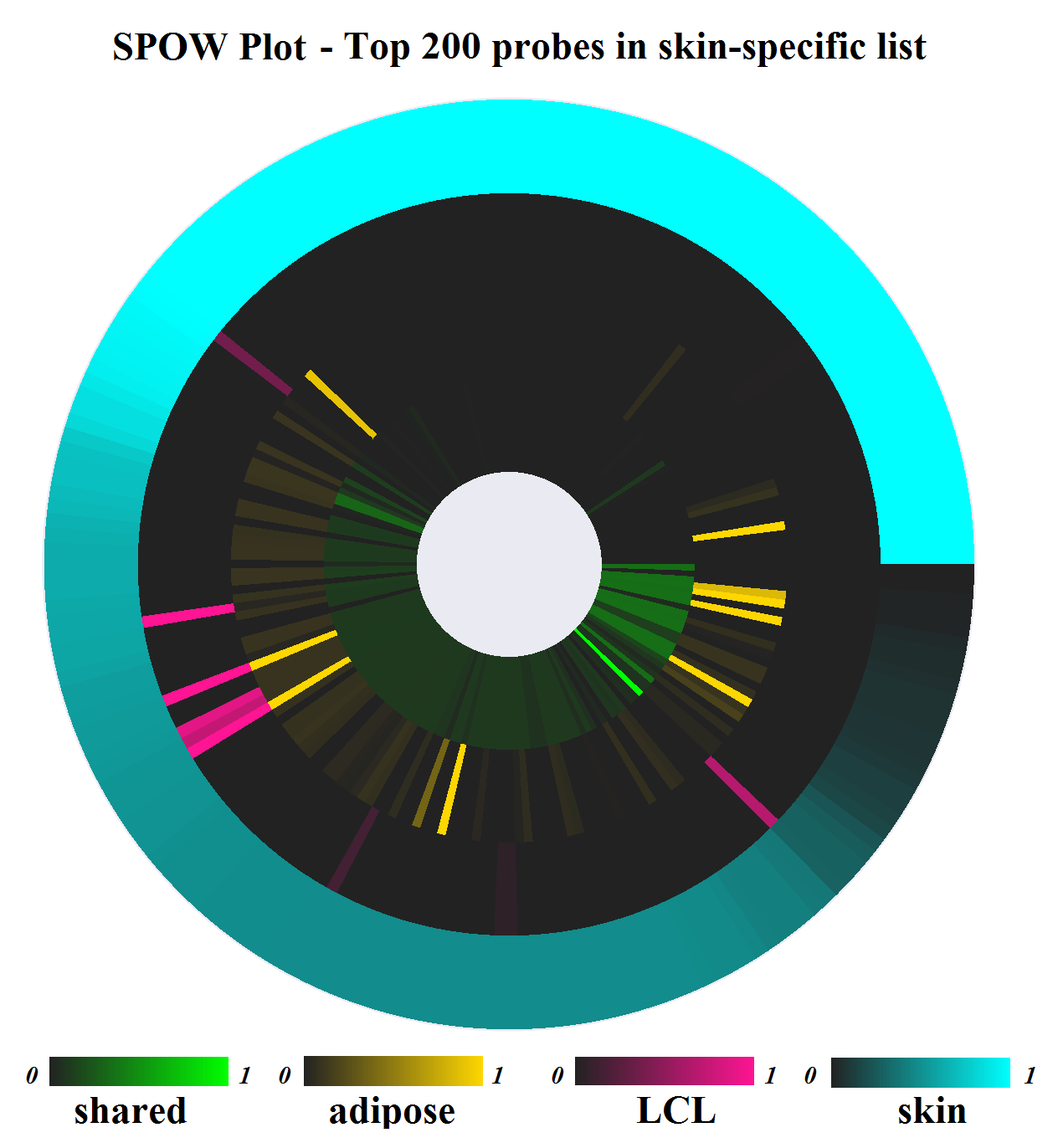}}
 \caption{TwinsUK study: SPOW plot $(d=r=3)$. The wheel contains the top 200 most frequently selected probes from the transformation matrix in the skin-specific component using sMVMF. We extract probes with bright colour in the skin-specific (cyan) ring and dark colours in the other rings.}
\label{skinSPOW}
\end{figure}

\begin{figure}[!tpb]
 \centerline{\includegraphics[scale=0.6]{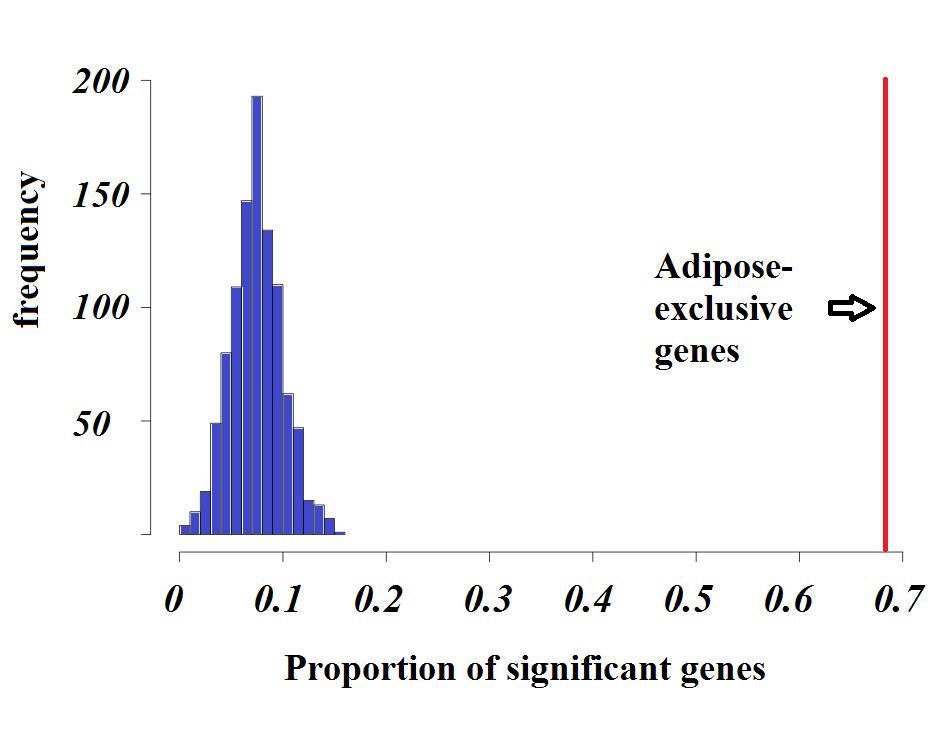}}
 \caption{Proportion of randomly chosen genes for which the corresponding gene expression shows a significant linear association with the methylation probe. The experiment consists of $1000$ random draws, and each draw involves $132$ randomly chosen expression probes, which are tested for linear association with the corresponding methylation profiles. We conclude that observing a proportion as large or larger  than $0.684$, which is what we obtained for our adipose-exclusive genes, is unlikely to happen by chance only.}
 \label{fig:fatMethyMean}
\end{figure} 

\end{appendices}

\end{document}